\documentclass{article}
\usepackage{spconf,amsmath,graphicx}
\usepackage{xcolor}
\usepackage{multirow}
\usepackage{amssymb}
\usepackage{subcaption}
\usepackage{tikz}
\usetikzlibrary{plotmarks}
\usepackage{pgfplots}
\usepackage{csvsimple}

\usepackage[justification=centering]{caption}
\usepackage{varwidth}

\pgfplotsset{compat=newest} % Allows to place the legend below plot
\usepgfplotslibrary{units} % Allows to enter the units nicely

\newenvironment{customlegend}[1][]{%
    \begingroup
    \csname pgfplots@init@cleared@structures\endcsname
    \pgfplotsset{#1}%
}{
    \csname pgfplots@createlegend\endcsname
    \endgroup
}

\def\addlegendimage{\csname pgfplots@addlegendimage\endcsname}

\newcommand\figref{{\bf Fig.}\ref}
\newcommand\tref{{\bf Table.}\ref}

\title{A Fully Convolutional Neural Network for Speech Enhancement}

\twoauthors
  {Se Rim Park}
	{Carnegie Mellon Univeristy\\%
	Pittsburgh, PA, 15213}
  {Jin Won Lee}
	{Qualcomm Research\\
	San Diego, CA, 92121}
\begin{document}
\ninept

%\tableofcontents

%\clearpage

\maketitle

\begin{abstract}
In hearing aids, the presence of babble noise degrades hearing intelligibility of human speech greatly.  However, removing the babble without creating artifacts in human speech is a challenging task in a low SNR environment. Here, we sought to solve the problem by finding a `mapping' between noisy speech spectra and clean speech spectra via supervised learning. Specifically, we propose using fully Convolutional Neural Networks, which consist of lesser number of parameters than fully connected networks. The proposed network, Redundant Convolutional Encoder Decoder (R-CED), demonstrates that a convolutional network can be 12 times smaller than a recurrent network and yet achieves better performance, which shows its applicability for an embedded system: the hearing aids. 

\end{abstract}


\section{Introduction}
\label{sec:intro}

Denoising speech signals has been a long standing problem. Decades of works showed feasible solutions which estimated the noise model and used it to recover noise-deducted speech \cite{boll1979suppression, lim1979enhancement, ephraim1984speech, scalart1996speech, ephraim1995signal}. Nonetheless, estimating the model for a babble noise, which is encountered when a crowd of people are talking, is still a challenging task.  

The presence of babble noise, however, degrades hearing intelligibility of human speech greatly. When babble noise dominates over speech, aforementioned methods often times will fail to find the correct noise model \cite{krishnamurthy2009babble}. If so, the noise-deduction will render distortion in speech, which creates discomforts to the users of hearing aids \cite{mccormack2013people}. 

Here, instead of explicitly modeling the babble noise, we focus on learning a `mapping' between noisy speech spectra and clean speech spectra, inspired by recent works on speech enhancement using neural networks \cite{han2014learning, xu2015regression, xia2013speech, osako2015complex}. However, the model size of Neural Networks easily exceeds several hundreds of megabytes, limiting its applicability for an embedded system.

On the other hand, Convolutional Neural Networks (CNN) typically consist of lesser number of parameters than FNNs and RNNs due to its weight sharing property. CNNs already proved its efficacy on extracting features in speech recognition \cite{abdel2014convolutional, amodei2015deep} or on eliminating noises in images \cite{mao2016image, he2015deep}. But upon our knowledge, CNNs have not been tested in speech enhancement.

In this paper, we attempted to find a `memory efficient' denoising algorithm for babble noise that creates minimal artifacts and that can be implemented in an embedded device: the hearing aid. Through experiments, we demonstrated that CNN can perform better than Feedforward Neural Networks (FNN) or Recurrent Neural Networks (RNN) with much smaller network size. A new network architecture, Redundant Convolutional Encoder Decoder (R-CED), is proposed, which extracts redundant representations of a noisy spectrum at the encoder and map it back to clean a spectrum at the decoder. This can be viewed as mapping the spectrum to  higher dimensions (e.g. kernel method), and projecting the features back to lower dimensions.
%Third, we found that extracting redundant information of the noisy spectrum is crucial for reconstructing back the clean signal as opposed to compressing the information which is normally applied in speech recognition work. This can be viewed as mapping the spectrogram to higher dimension (e.g. kernel method), and projecting the features back to the original space. 

The paper is organized as follows. In section \ref{sec:map}, a formal definition of the problem is stated. In section \ref{sec:conv}, the fully convolutional network architectures are presented including the proposed R-CED network. In section \ref{sec:exp}, the experimental methods are provided. In section \ref{sec:config}, description of the experiments and the corresponding network configurations are provided. In section \ref{sec:result}, the results are discussed, and in section \ref{sec:con}, we end with conclusion of this study.

% Literature Review 4. Image de-noising & Speech recognition using CNN
%On the other hand, CNN usually requires much smaller number of parameters to be trained per layer due to its weight sharing property which make it more powerful for modeling a function by stacking deep \cite{szegedy2015goting}

\begin{figure}[t!]
\centering
\includegraphics[width=\columnwidth]{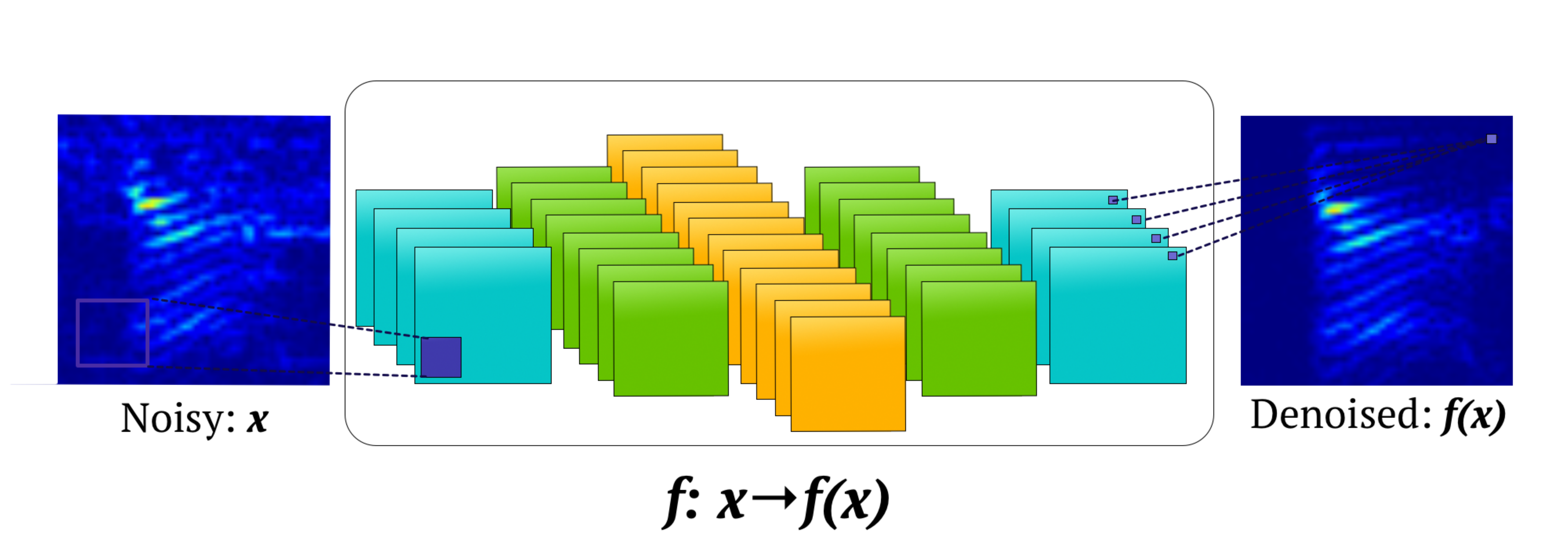}
\caption{Speech Enhancement Using a CNN}
\label{fig:map}
\end{figure}

\begin{figure*}
\begin{minipage}[t]{0.58\textwidth}
\includegraphics[width=\textwidth]{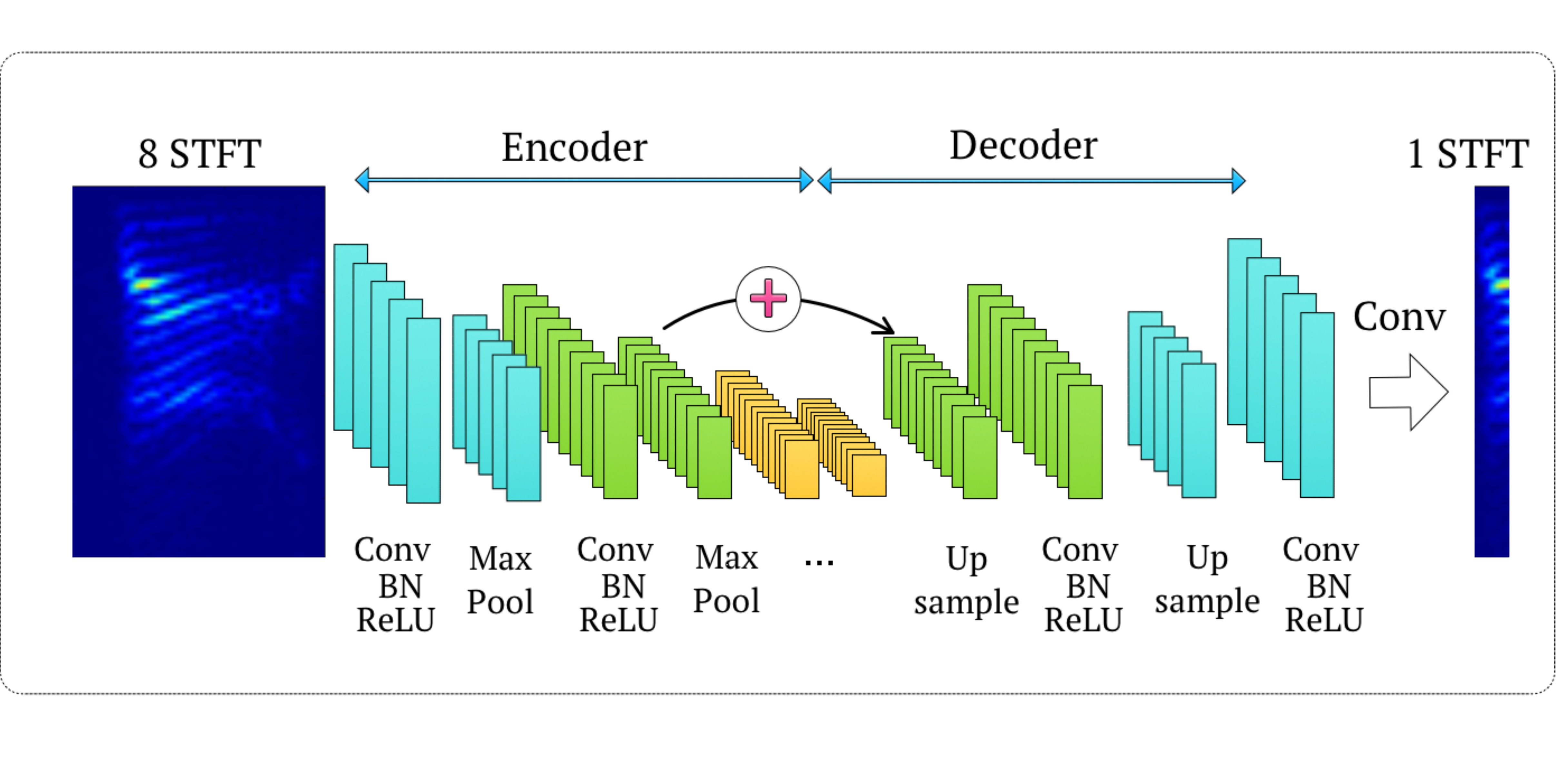}
\caption{Modified Convolutional Encoder-Decoder Network (CED)}
\label{fig:CED}
\end{minipage}
%\hspace{-1cm}
\begin{minipage}[t]{0.45\textwidth}
%\vspace{0.1cm}
\includegraphics[width=\textwidth]{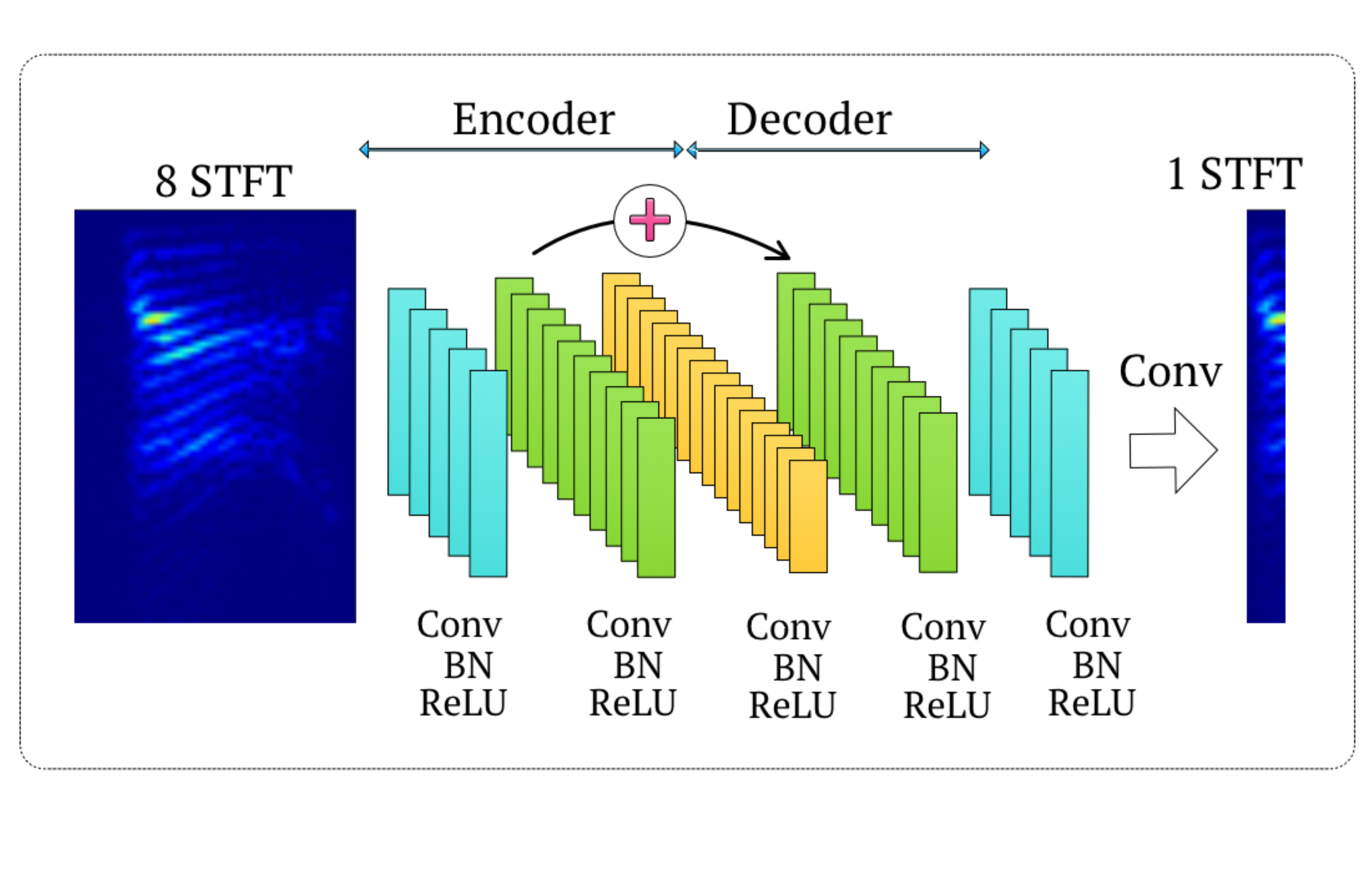}
\caption{Proposed Redundant CED (R-CED)}
\label{fig:RCED}
\end{minipage}
\end{figure*}

\section{Problem Statement}
\label{sec:map}
Given a segment of noisy spectra $\{{\bf x}_t\}_{t=1}^T$ and clean spectra $\{{\bf y}_t\}_{t=1}^T$, our aim is to learn a mapping $f$ which generates a segment of `denoised' spectra $\{f({\bf x}_t)\}_{t=1}^T$ that approximate the clean spectra in the $\ell_2$ norm, e.g. 
% Formally, $f$ minimizes the sum of $\ell_2$ errors between the clean spectra ${\bf y}_t$ the denoised spectra $f({\bf x}_t)$, e.g.
\begin{align}
\min \sum_{t=1}^T ||{\bf y}_t - f({\bf x}_t)||_2^2.
\label{eq:obj_rnn}
\end{align}
Specifically, we formulate $f$ using a Neural Network (see \figref{fig:map}). If $f$ is a recurrent type network, the temporal behavior of input spectra is already addressed by the network, and hence objective \eqref{eq:obj_rnn} suffices. On the other hand, for a convolutional type network, the past $n_T$ noisy spectra $\left\{ {\bf x}_i \right\}_{i=t-n_T+1}^t$ are considered to denoise the current spectra, e.g.
\begin{align}
 \sum_{t=1}^T ||{\bf y}_t - f({\bf x}_{t-n_T+1, \cdots} , {\bf x}_t)||_2^2.
\end{align}
We set $n_T= 8$. Hence, input spectra to the network is equivalent to about 100ms of speech segment, whereas the output spectra of the network is of duration 32ms (see \figref{fig:CED}, \figref{fig:RCED}). 
 
\section{Convolutional Network Architectures}
\label{sec:conv}

\subsection{Convolutional Encoder-Decoder Network (CED)}
Convolutional Encoder-Decoder (CED) network proposed in \cite{vincent2010stacked} consists of symmetric encoding layers and decoding layers (see \figref{fig:CED}, each block represents a feature). Encoder consists of repetitions of a convolution, batch-normalization \cite{ioffe2015batch}, max-pooling, and an ReLU \cite{nair2010rectified} activation layer. Decoder consists of repetitions of a convolution, batch-normalization, and an up-sampling layer. Typically, CED compresses the features along the encoder, and then reconstructs the features along the decoder. In our problem, the orignal Softmax layer at the last layer is modified to a convolution layer, to make CED fully convolutional network.

\subsection{Redundant CED Network (R-CED)}
Here, we propose an alternative convolutional network architecture, namely Redundant Convolutional Encoder-Decoder (R-CED) network. R-CED consists of repetitions of a convolution, batch-normalization, and a ReLU activation layer (see \figref{fig:RCED}, each block represents a feature). No pooling layer is present, and thus no upsampling layer is required. As opposite to CED, R-CED encodes the features into higher dimension along the encoder and achieves compression along the decoder. The number of filters are kept symmetric: at the encoder, the number of filters are gradually increased, and at the decoder, the number of filters are gradually decreased. The last layer is a convolution layer, which makes R-CED a fully convolutional network.

{\bf Cascaded R-CED Network (CR-CED):}
Cascaded Redundant Convolutional Encoder-Decoder Network (CR-CED) is a variation of R-CED network. It consists of repetitions of R-CED Networks. Compared to the R-CED with the same network size (= the same number of parameters), CR-CED achieves better performance with less convergence time.

\subsection{Bypass Connections}
For CED, R-CED, and CR-CED, bypass connections are added to the network to facilitate the optimization in the training phase and improve performance. Between two different bypass schemes --- skip connections in \cite{mao2016image} and residual connections in \cite{he2015deep} --- we chose to use skip connections in \cite{mao2016image} which is more suitable for symmetric encoder-decoder design. Bypass connections are illustrated in \figref{fig:CED} and \figref{fig:RCED} as an `addition' operation symbol with an arrow. Bypass connections are added every other layer.

\subsection{1-Dim Convolution Operation for Convolution Layers}
\begin{figure}[t!]
\centering
\hspace{-0.5cm}
\begin{minipage}[t]{0.24\textwidth}
\includegraphics[width=\textwidth]{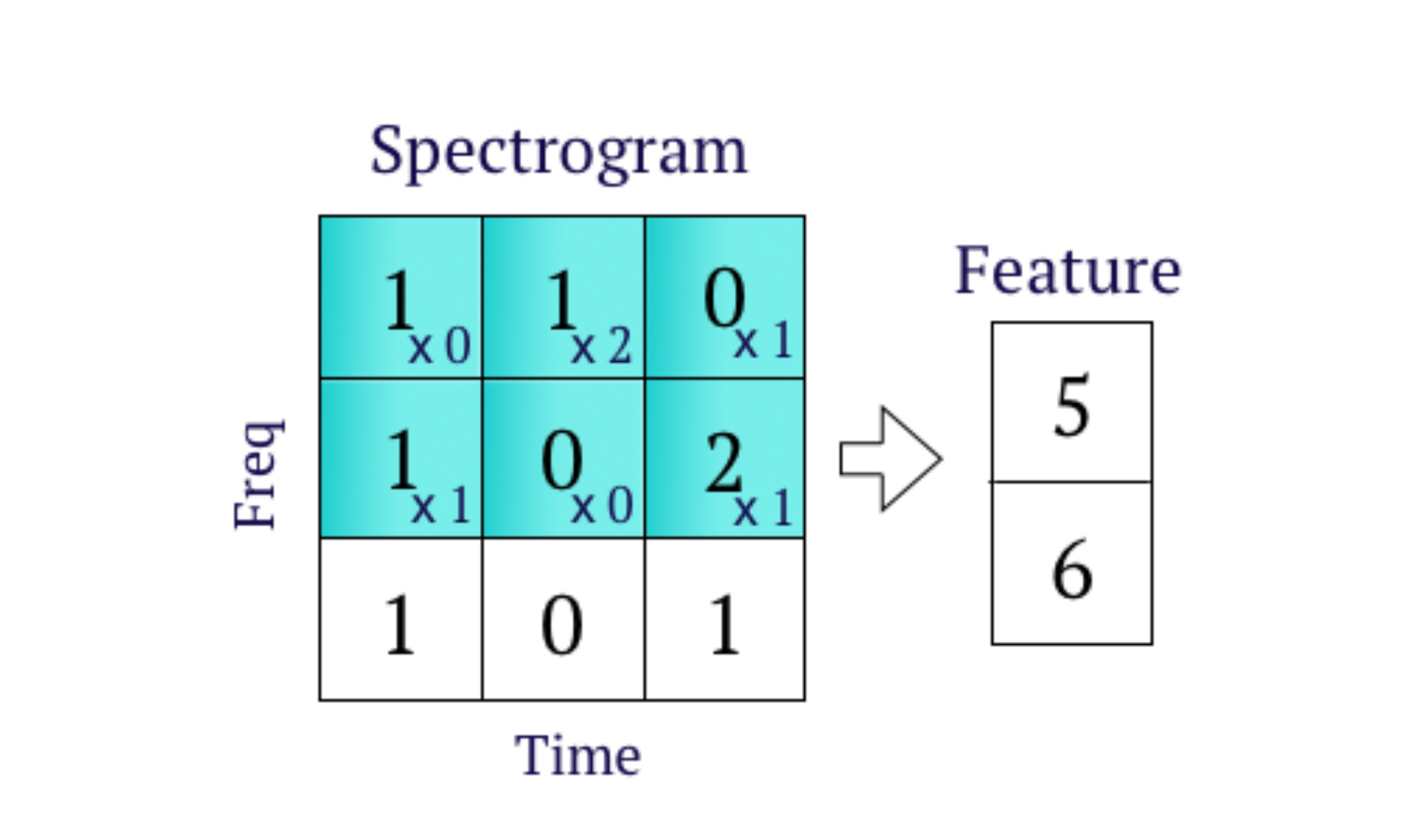}
\caption{1-d Convolution}
\label{fig:1dimConvSpec}
\end{minipage}
\begin{minipage}[t]{0.24\textwidth}
\includegraphics[width=\textwidth]{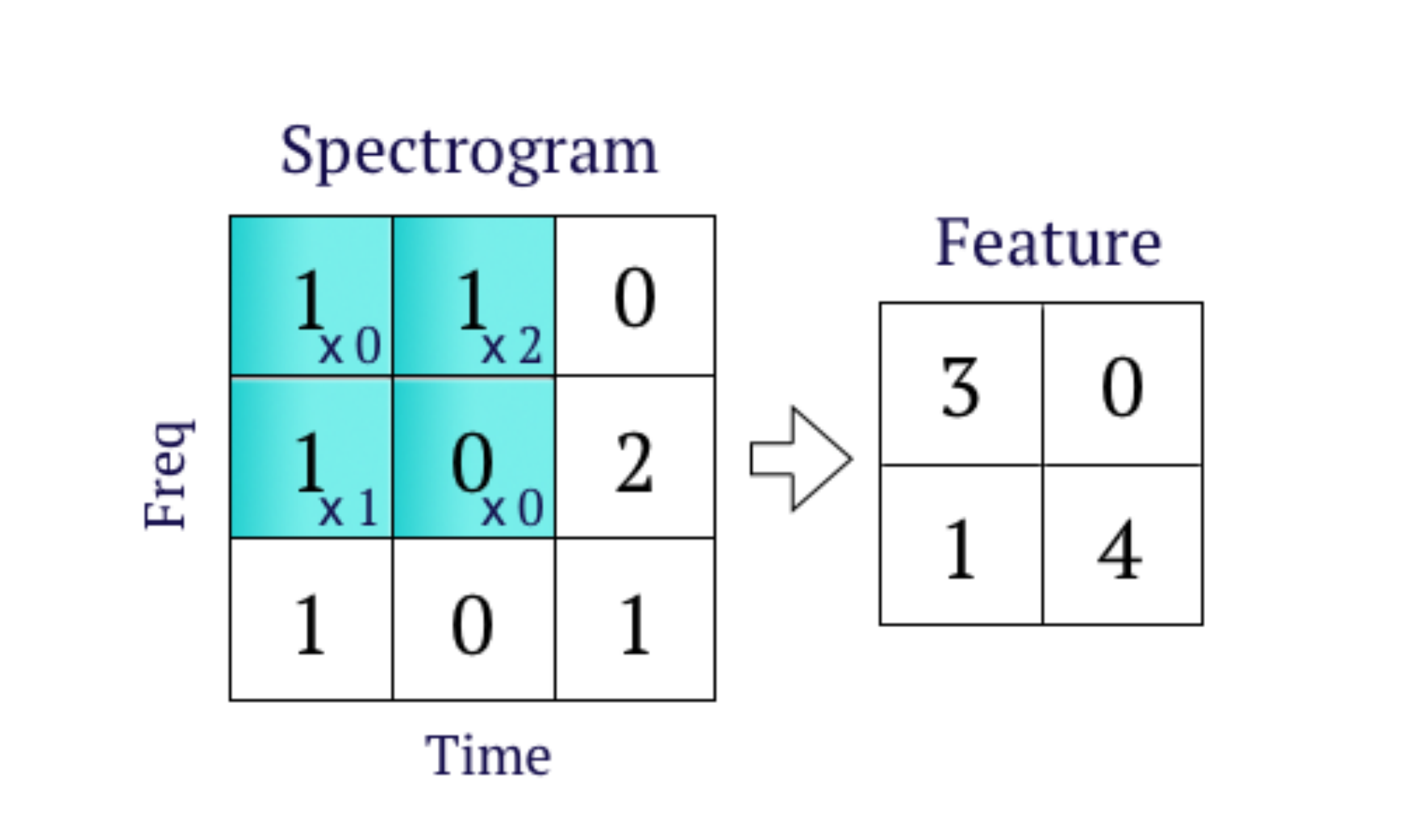}
\caption{2-d Convolution}
\label{fig:2dimConv}
\end{minipage}
\end{figure}

At all convolution layers throughout the paper, convolution was performed only in 1 direction (see \figref{fig:1dimConvSpec}). In \figref{fig:1dimConvSpec}, the input ($3 \times 3$ white matrix) and the filter ($2 \times 3$ blue matrix) has the same dimension in time axis, and convolution is performed in frequency axis. We found this more efficient than 2-dim convolution (see \figref{fig:2dimConv}) for our input spectra ($129 \times 8$).

\begin{figure*}[t!]
\centering
\begin{minipage}[t]{0.32\textwidth}
\includegraphics[width=\textwidth]{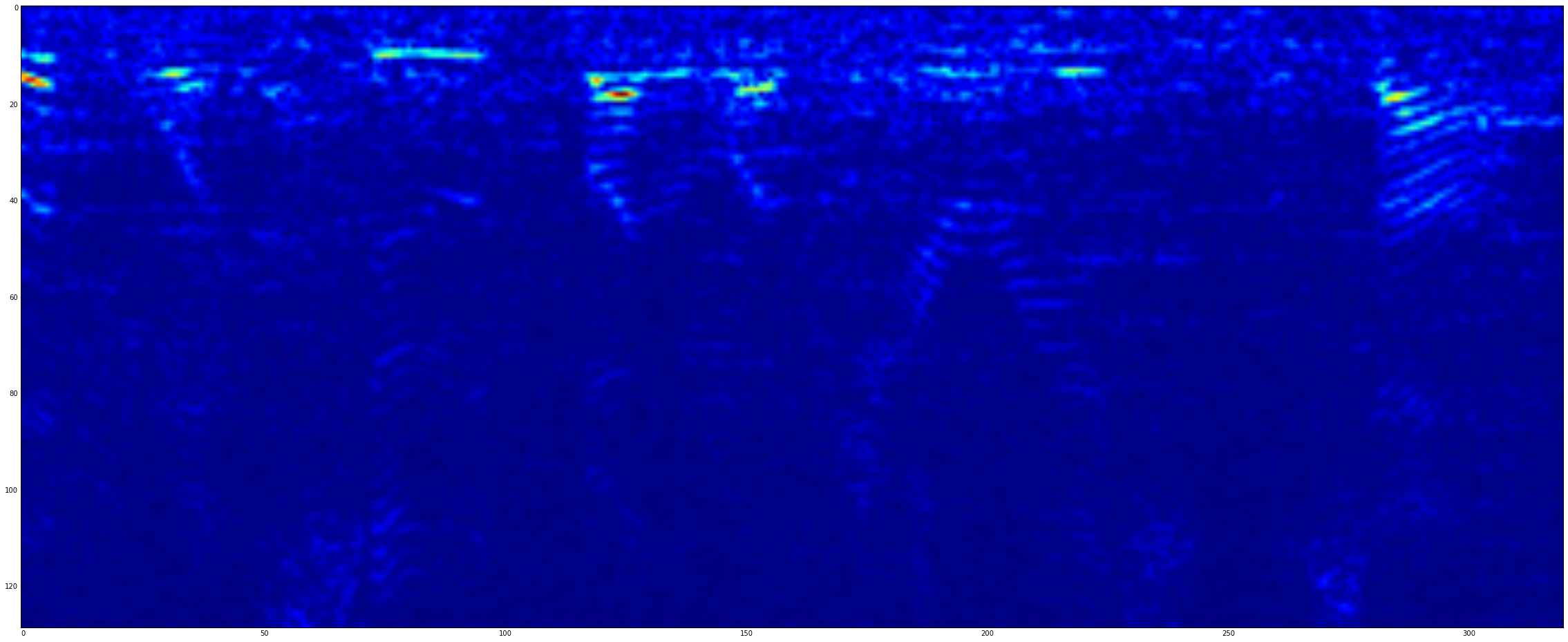}
\caption{Noisy Spectrogram}
\label{fig:noisy_mag}
\end{minipage}
\begin{minipage}[t]{0.32\textwidth}
\includegraphics[width=\textwidth]{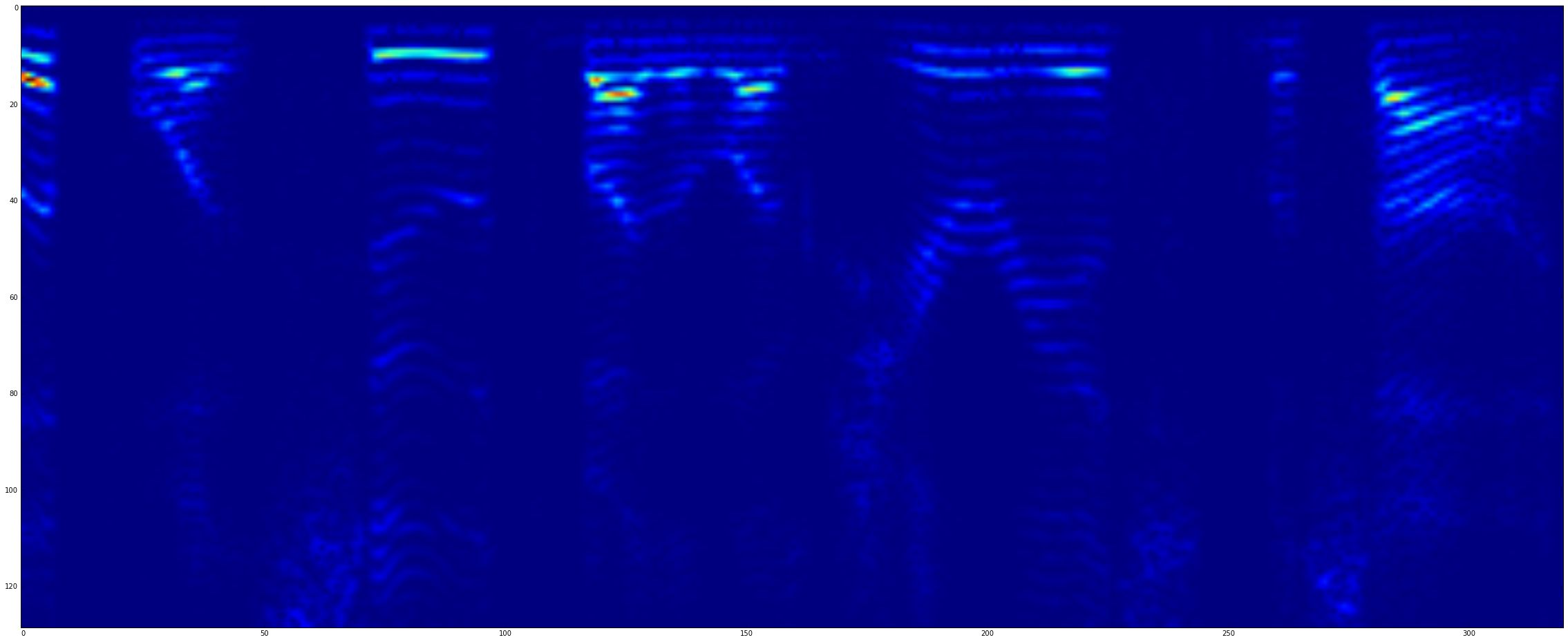}
\caption{Clean Spectrogram}
\label{fig:clean_mag}
\end{minipage}
\begin{minipage}[t]{0.32\textwidth}
\includegraphics[width=\textwidth]{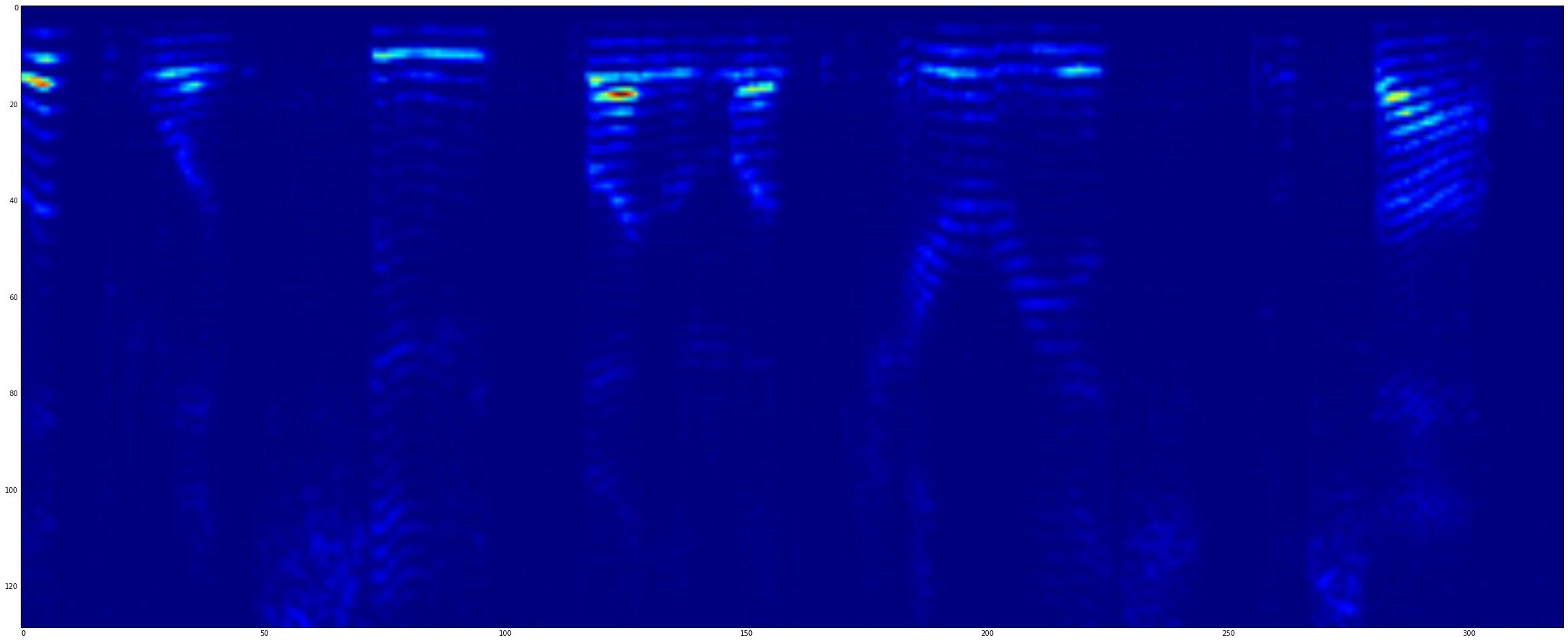}
\caption{Denoised Spectrogram}
\label{fig:denoised_mag}
\end{minipage}
\end{figure*}

\begin{table*}[h!]
%\begin{minipage}[b]{0.63\textwidth}
\begin{tabular}{|c|c|c|c|}
\hline
 & {\bf Layer Configuration }& {\bf Number of Filters }& {\bf Filter Width}\\
 \hline
\multirow{1}{*}{ {\bf CED, CED /w Bypass } } &Encoder: (Conv, BN, ReLU, Pool) $\times$ 5, & 12-16-20-24-32- & 13-11-9-7-5-\\
(11 Conv) &Decoder: (Conv, BN, ReLU, Upsample) $\times$ 5,  Conv. & 24-20-16-12-8-1 & 7-9-11-13-8-129\\
 \hline
\multirow{1}{*}{{\bf R-CED, R-CED /w Bypass}} & \multirow{2}{*}{(Conv, ReLU, BN) $\times$ 9, Conv.} &  12-16-20-24-32- & 13-11-9-7-7-\\
 (10 Conv)& & 24-20-16-12-1 & 7-9-11-13-129\\
 \hline
\multirow{1}{*}{{\bf R-CED, R-CED /w Bypass}} & \multirow{2}{*}{(Conv, ReLU, BN) $\times$ 15, Conv.} & 10-12-14-15-19-21-23-25- & 11-7-5-5-5-5-7-11- \\
 (16 Conv)& & 23-21-19-15-14-12-10-1 &  7-5-5-5-5-7-11-129\\
\hline
{\bf CR-CED (16 Conv)} & (Conv, ReLU, BN) $\times$ 15, Conv. & (18-30-8) $\times$ 5, 1 & (9-5-9) $\times$ 5, 129 \\
\hline
\end{tabular}
\caption{Network Configurations for CNNs: CED vs. R-CED vs. CR-CED}
\label{table:config_cnn}
%\end{minipage}
\end{table*}

\section{Experimental Methods}
\label{sec:exp}

\subsection{Preprocessing}

\hspace{0.4cm} {\bf Dataset:} The experiment was conducted on the TIMIT database \cite{garofolo1993darpa} and 27 different types of noise clips were collected from freely available online resource \cite{akkermans2011freesound}. The noise are mostly babble, but includes different types of noise like instrumental sounds. Both data in the training set (4620 utterances) and the testing set (200 utterances) were added with one of 27 noise clips at 0dB SNR.  After all feature transformation steps were completed, 20\% of the training features were assigned as the validation set. 

{\bf Feature Transformation:}
The audio signals were down-sampled to 8kHz, and the silent frames were removed from the signal. The spectral vectors were computed using a 256-point Short Time Fourier Transform (32ms hamming window) with a window shift of 64-point (8ms). The frequency resolution was 31.25 Hz (=4kHz/128) per each frequency bin. 256-point STFT magnitude vectors were reduced to 129-point by removing the symmetric half. For FNN/RNN, the input feature consisted of a noisy STFT magnitude vector (size: 129$\times$1, duration: 32ms). For CNN, the input feature consisted of 8 consecutive noisy STFT magnitude vectors (size: $129\times8$,  duration: 100ms). Both input features were standardized to have zero mean and and unit variance.

{\bf Phase Aware Scaling:}
To avoid extreme differences (more than 45 degree) between the noisy and clean phase, the clean spectral magnitude was encoded as similar to \cite{mowlaee2013iterative}:
\[
{\bf s}_{\text{phase aware}} = {\bf s}_{\text{clean}}\cos(\theta_{\text{clean}} - \theta_{\text{noisy}}).
\]
Besides, spectral phase was not used in the training phase. At reconstruction, noisy spectral phase was used instead to perform inverse STFT and recover human speech. However, because human ear is not susceptible to phase difference smaller than 45 degree, the distortion was negligible. Through `phase aware scaling', the phase mismatch be smaller than 45 degree, and  For all networks, the output feature consisted of  a `phase aware' magnitude vector (size: 129$\times$1, duration: 32ms), and were standardized to have zero mean and and unit variance. 

\subsection{Optimization}
Fully connected and convolution layer weights were initialized as in \cite{glorot2010understanding} and recurrent layer weights were initialized as in \cite{le2015simple}. Fully connected and recurrent layer weights were pretrained from the networks of smaller depth with same number of nodes. Convolution layers were trained from scratch, with the aid of batch normalization layer \cite{ioffe2015batch} added after each convolution layer \footnote{We note that adding BN layers on both FNN and RNN did not improve convergence nor performance for this experiment.}. All networks were trained using back propagation with gradient descent optimization using Adam \cite{Adam} with a mini-batch size of 64. The learning rate started from $lr =0.0015$ with $\beta_1 = 0.9$, $\beta_2 = 0.999$, and $\epsilon = 1.0e^{-8}$. When the validation loss didn't decrease for more than 4 epochs, learning rate was decreased to $lr/2, lr/3, lr/4$, subsequently. The training was repeated once more for FNN and RNN with $\ell_2$ regularization ($\lambda=10^{-5}$) which slightly improved the performance.

\subsection{Evaluation Metric}
Signal to Distortion Ration (SDR) \cite{ vincent2006performance} was used to measure the amount of $\ell_2$ error present between clean and denoised speech:
\[
SDR := 10 \log_{10} \frac{ || {\bf y } ||^2 } { || f({\bf x} ) - {\bf y}  || ^2 }.
\]
SDR is inversely associated with the objective function presented in \eqref{eq:obj_rnn}. In addition, Short time Objective Intelligibility (STOI) \cite{taal2010short} and Perceptual Evaluation of Speech Distortion (PESQ) \cite{rix2001perceptual} --- both assume that human perception has short term memory and hence the error is measured nonlinearly in time of interest--- were used to measure the subjective quality of listening.

\section{Experimental Setup}
\label{sec:config}

\subsection{Test 1: FNN vs. RNN vs. CNN}
The first experiment compared CNN with FNN and RNN to demonstrate how feasible it is to use CNN for speech enhancement. Network configurations (e.g. number of nodes, number of layers) that yielded the best performance for each network are summarized in \tref{table:config_nn}. The best FNN and RNN architectures have 4 fully connected (FC) layers whereas CNN has 16 convolutional layers. 
%For FNN and RNN, weights were pretrained from the networks of the smaller depth with the same number of nodes. CNN was trained from scratch, with Batch Normalization (BN) Layer. 
\begin{table}[b!]
\begin{tabular}{|c|c|c|}
 \hline
 & \# of Layers & \# of Nodes \\
 \hline
\multirow{2}{*}{\bf FNN} & (FC, ReLU) $\times$ 3, & \multirow{2}{*}{1024-1024-1024}\\
&  FC&\\
\hline
\multirow{2}{*}{\bf RNN} & (Recurrent, ReLU) $\times$ 3, & \multirow{2}{*}{256-256-256}\\
&  FC &\\
\hline
\multirow{1}{*}{\bf CNN} & (Conv, ReLU, BN) $\times$ 15,& 10-12-14-15-19-21-23-25\\
(R-CED)& Conv   &23-21-19-15-14-12-10-1\\
\hline
\end{tabular}
\caption{Network Config. for FNN vs. RNN vs. CNN}
\label{table:config_nn}
\end{table}

\subsection{Test 2: CED vs. R-CED}
In the second experiment, R-CED was compared to CED. For fair comparison, the total number of parameters were fixed to 33,000 (roughly 132MB of memory) while the depth of the network was fixed to 10 convolution layers. The filter width per each layer is determined accordingly such that i) the symmetric encoder-decoder structure is maintained, ii) the number of parameters are gradually increased and decreased, iii) the `frequency coverage' is equal for both network. Here, `frequency coverage' refers to how much nearby frequency bins at the input are used to reconstruct a single frequency bin at the output. We made sure that both networks utilize the same amount of frequency bins to reconstruct a single frequency bin. The configurations for Test 2 is summarized at top two rows of \tref{table:config_cnn}.

\subsection{Test 3: Finding the Best R-CED Performance}
In the third experiment, we tested how far the performance can be improved using the R-CED network. The R-CED and CR-CED network with skip connections of various network size and depth are compared. The network size (the number of parameters) considered are 33K (132MB memory) and 100K (400MB memory). The depth of the network considered are 10, 16, and 20 convolution layers. Bottom 3 rows in \tref{table:config_cnn} summarize the network configurations for Test 3. 

\section{Results}
\label{sec:result}
\subsection{Test 1: FNN vs. RNN vs. CNN}
\figref{fig:perf_dnn} illustrates the denoising performance of FNN, RNN and CNN (left), and the corresponding network size (=the number of parameters, right). All networks exhibited similar performance based on both subjective (STOI, PESQ) and objective quality (SDR) measure. On the other hand, the model size of CNN was about 68 times smaller than that of FNN and about 12 times smaller than RNN. We note that FNN and RNN were optimized to have the smallest network architectures. This experiment validates that CNN requires far lesser number of parameters per layer due to its weight sharing property, and yet can achieve similar or better performance compared to FNN and RNN. 33,000 parameters for CNN are roughly 132MB of memory which can be implemented in an embedded system. Refer to \figref{fig:noisy_mag}, \figref{fig:clean_mag}, \figref{fig:denoised_mag} for the example of noisy spectrogram, clean spectrogram, and denoised spectrogram from CNN respectively.

\subsection{Test 2: CED vs. R-CED}
The denoising performance of CED and R-CED are shown in \figref{fig:perf} with first 4 bars. The R-CED with skip connections showed the best performance, whereas the CED without skip connections showed the worst performance. Regardless of the presence of skip connections, R-CED yielded better results than CED. 

The effect of the skip connection was prominent in CED (5.96 to 7.92). This implies that the decoder itself could not reconstruct the `lost' information compressed at the encoder, unless the `lost' information was provided by the skip connections. In addition, the resulting speech from CED sounded artificial and mechanical. This confirms that the decoder could not reconstruct what is necessary for audios to sound like human speech. 

On the other hand, the effect of the skip connection was not that notable in R-CED (8.07 to 8.19). This is because R-CED  rather expands than compresses the input at the encoder, which can be viewed as mapping a spectrum to higher dimension (e.g. kernel method). By generating redundant representations of important features at the encoder, and removing unwanted features at the decoder, the speech quality was effectively enhanced. 

\subsection{Test 3: Finding the Best R-CED Network}
The denoising performance of R-CED of various network size and depth are presented in \figref{fig:perf} with the last five bars. A few interesting observations are i) the network size was the most dominant factor that was associated with the network performance, ii) the network depth was secondary, iii) CR-CED with skip connection yielded the best performance when other conditions were kept same (16 convolution layers, 33K parameters).

\definecolor{color1}{RGB}{179,205,227}
\definecolor{color2}{RGB}{140,150,198}
\definecolor{color3}{RGB}{136,86,167}
\definecolor{color4}{RGB}{129,15,124}

\pgfplotscreateplotcyclelist{param-colors}{
{color1},
{color2},
{color3},
}

\begin{figure}[t!]
\centering
   \begin{tikzpicture}
   \hspace{-1cm}
    \begin{customlegend}[legend columns =3, legend style={column sep=1ex}, 
legend entries={STOI, PESQ, SDR, \#Params }, 
    	legend image code/.code={%
                    \draw[] (0cm,-0.1cm) rectangle (0.3cm,0.1cm);
                } ]
    \addlegendimage{black, fill =color1}
     \addlegendimage{black, fill = color2}
    \addlegendimage{black, fill =color3}
%        \addlegendimage{area legend, brown, fill =black!30}
    \end{customlegend}
       \end{tikzpicture}

  \begin{tikzpicture}
      \begin{axis}[
      compat=newest,
      	   ybar,
	   x=1.5cm,
	   bar width = 0.3cm,
	   enlarge x limits ={abs=0.5},
	   enlarge y limits ={upper, abs value=4},
          width={10cm}, % Scale the plot to \linewidth
          height = 4cm,
          grid=major, % Display a grid
          grid style={dashed,gray!30}, % Set the style
  %        xlabel=Network Types, % Set the labels
          ylabel=Performance,
          ymin = 0,
          xtick={0,1,2},
          xticklabels={FNN,RNN,CNN},
           nodes near coords,
           every node near coord/.append style={rotate=90, anchor=west},
%    	    nodes near coords align={vertical},
%           legend style={at={(0.5,-0.2)},anchor=north}, % Put the legend below the plot
%           legend={STOI, PESQ, SDR},
           x tick label style={rotate=45, anchor=north east, inner sep=0mm, font=\bfseries},
        ]
%        \addplot [ybar, fill= color1brown!50] table [y = STOI,  col sep=comma]{result/FNN_RNN_CNN_bar.csv};
%        \addplot [ybar, fill=blue!50] table [y = PESQ,  col sep=comma]{result/FNN_RNN_CNN_bar.csv};
%        \addplot [ybar, fill=red!50] table [y = SDR,  col sep=comma]{result/FNN_RNN_CNN_bar.csv};
        \addplot [ybar, fill= color1] table [y = STOI,  col sep=comma]{result/FNN_RNN_CNN_bar.csv};
        \addplot [ybar, fill= color2] table [y = PESQ,  col sep=comma]{result/FNN_RNN_CNN_bar.csv};
        \addplot [ybar, fill= color3] table [y = SDR,  col sep=comma]{result/FNN_RNN_CNN_bar.csv};
%        \addplot [ybar, fill=black!50] table [x = Entry, y = Params-norm, col sep=comma]{result/FNN_RNN_CNN_bar.csv};
      \end{axis}
    \end{tikzpicture}
\hfill
   \begin{tikzpicture}
      \begin{axis}[
            compat=newest,
      	   ybar,
	   x=0.5cm,
	   bar width = 0.3cm,
	   enlarge x limits ={abs=0.6},
          enlarge y limits ={upper, abs value=1.2},
          width={10cm}, % Scale the plot to \linewidth
          height = 4cm,
          grid=major, % Display a grid
          grid style={dashed,gray!30}, % Set the style
%          xlabel=Network Types, % Set the labels
         ylabel=\# of Params (M),
         ymin = 0,
           xtick={0,1,2},
          xticklabels={FNN,RNN,CNN},
           nodes near coords,
           every node near coord/.append style={rotate=90, anchor=west, },
         x tick label style={rotate=45, anchor=north east, inner sep=0mm, font=\bfseries},
        ]
        \addplot [ybar, fill= color2] table [y = Params(M),  col sep=comma]{result/FNN_RNN_CNN_bar.csv};
      \end{axis}
    \end{tikzpicture}
     \caption{A Comparison of Denoising Performance and the Network Size for FNN vs. RNN vs. CNN}
   \label{fig:perf_dnn}
   \end{figure}
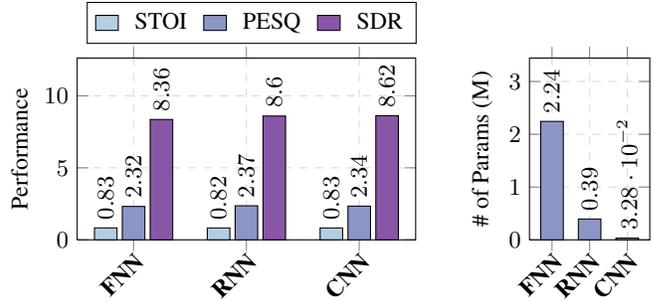

\definecolor{L10}{RGB}{203,201,226}
%\definecolor{L16}{RGB}{158,154,200}
\definecolor{L16}{RGB}{117,107,177}
\definecolor{L20}{RGB}{84,39,143}

\begin{figure}[t!]
\centering
    \begin{tikzpicture}
      \begin{axis}[
            compat=newest,
%      	   ybar =- 0.5cm,
	   bar width = 0.3cm,
          enlarge y limits ={upper, abs value =2},
          ymin = 5,
          width=\linewidth, % Scale the plot to \linewidth
          height = 4cm,
          grid=major, % Display a grid
          grid style={dashed,gray!30}, % Set the style
          ylabel=Performance (SDR),
         xtick={0,1,2,3,4,5,6,7},
        xticklabels={CED, CED (skip), R-CED, R-CED (skip), R-CED, R-CED (skip),   CR-CED (skip), R-CED (skip)},
         x tick label style={rotate=45, anchor=north east, inner sep=0mm, font=\bfseries},
           nodes near coords,
           every node near coord/.append style={rotate=90, anchor=west, },
           cycle list ={
           	{fill=L10, draw=black!40}, 
	        {fill=L10, draw=black!40}, 
		{fill=L10, draw=black!40}, 
		{fill=L10, draw=black!40}, 
		{fill=L16, draw=black!40}, 
		{fill=L16, draw=black!40}, 
		{fill=L16, draw=black!40}, 
		{fill=L20, draw=black!40}}
        ]
\addplot [ybar, fill=L10, draw=black!50] coordinates{( 0, 5.96)};
\addplot [ybar, fill=L10, draw=black!50] coordinates{( 1, 7.92)};
\addplot [ybar, fill=L10, draw=black!50] coordinates{( 2, 8.07)};
\addplot [ybar, fill=L10, draw=black!50] coordinates{( 3, 8.19)};
\addplot [ybar, fill=L16, draw=black!50] coordinates{( 4, 8.41)};
\addplot [ybar, fill=L16, draw=black!50] coordinates{( 5, 8.62)};
\addplot [ybar, fill=L16, draw=black!50] coordinates{( 6, 8.73)};
\addplot [ybar, fill=L20, draw=black!50] coordinates{( 7, 8.79)};
      \end{axis}
    \end{tikzpicture}

    \begin{tikzpicture}
    \hspace{0.2cm}
    \vspace{0.2cm}
    \begin{customlegend}[legend columns =3, legend style={column sep=1ex}, legend entries={10Conv(33K), 16Conv(33K), 20Conv(100K)}, 
    	legend image code/.code={%
                    \draw[] (0cm,-0.1cm) rectangle (0.3cm,0.1cm);
                } ]
    \addlegendimage{fill=L10, draw=black!50}
   \addlegendimage{ fill=L16, draw=black!50}
     \addlegendimage{fill=L20, draw=black!50}
    \end{customlegend}
\end{tikzpicture}

   \caption{A Comparison of Denoising Performance and the Model Size for different CNN Archtectures}
\label{fig:perf}
\end{figure}
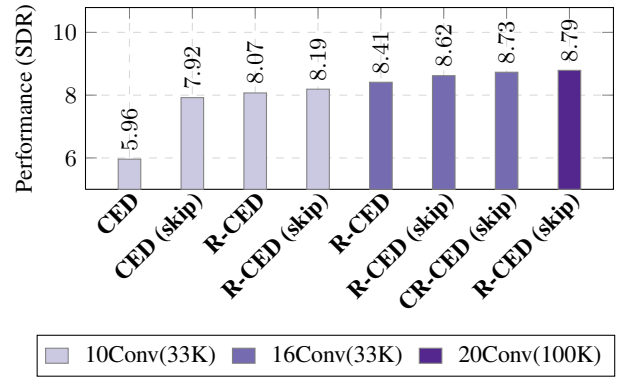

\section{Conclusion}
\label{sec:con}
In this paper, we aimed to find a memory efficient denoising method that can be implemented in an embedded system. Inspired by past success of FNN and RNN, we hypothesized that CNN can effectively denoise speech with smaller network size according to its weight sharing property. We set up an experiment to denoise human speech from babble noise which is a major discomfort to the users of hearing aids. Through experiments, we demonstrated that CNN can yield similar or better performance with much lesser number of model parameters compared to FNN and RNN. Also, we proposed a new fully convolutional network architecture R-CED, and showed its efficacy in speech enhancement. We observed that the success of R-CED is associated with the increasing dimension of the feature space along the encoder and decreasing dimension along the decoder. We expect that R-CED can be also applied to other interesting domains. Our future work will include pruning the R-CED to minimize the operation count of convolution. 

\bibliographystyle{IEEEbib}
\bibliography{paper_v4}

\begin{thebibliography}{10}

\bibitem{boll1979suppression}
Steven Boll,
\newblock ``Suppression of acoustic noise in speech using spectral
  subtraction,''
\newblock {\em IEEE Transactions on acoustics, speech, and signal processing},
  vol. 27, no. 2, pp. 113--120, 1979.

\bibitem{lim1979enhancement}
Jae~S Lim and Alan~V Oppenheim,
\newblock ``Enhancement and bandwidth compression of noisy speech,''
\newblock {\em Proceedings of the IEEE}, vol. 67, no. 12, pp. 1586--1604, 1979.

\bibitem{ephraim1984speech}
Yariv Ephraim and David Malah,
\newblock ``Speech enhancement using a minimum-mean square error short-time
  spectral amplitude estimator,''
\newblock {\em IEEE Transactions on Acoustics, Speech, and Signal Processing},
  vol. 32, no. 6, pp. 1109--1121, 1984.

\bibitem{scalart1996speech}
Pascal Scalart et~al.,
\newblock ``Speech enhancement based on a priori signal to noise estimation,''
\newblock in {\em Acoustics, Speech, and Signal Processing, 1996. ICASSP-96.
  Conference Proceedings., 1996 IEEE International Conference on}. IEEE, 1996,
  vol.~2, pp. 629--632.

\bibitem{ephraim1995signal}
Yariv Ephraim and Harry~L Van~Trees,
\newblock ``A signal subspace approach for speech enhancement,''
\newblock {\em IEEE Transactions on speech and audio processing}, vol. 3, no.
  4, pp. 251--266, 1995.

\bibitem{krishnamurthy2009babble}
Nitish Krishnamurthy and John~HL Hansen,
\newblock ``Babble noise: modeling, analysis, and applications,''
\newblock {\em IEEE transactions on audio, speech, and language processing},
  vol. 17, no. 7, pp. 1394--1407, 2009.

\bibitem{mccormack2013people}
Abby McCormack and Heather Fortnum,
\newblock ``Why do people fitted with hearing aids not wear them?,''
\newblock {\em International Journal of Audiology}, vol. 52, no. 5, pp.
  360--368, 2013.

\bibitem{han2014learning}
Kun Han, Yuxuan Wang, and DeLiang Wang,
\newblock ``Learning spectral mapping for speech dereverberation,''
\newblock in {\em 2014 IEEE International Conference on Acoustics, Speech and
  Signal Processing (ICASSP)}. IEEE, 2014, pp. 4628--4632.

\bibitem{xu2015regression}
Yong Xu, Jun Du, Li-Rong Dai, and Chin-Hui Lee,
\newblock ``A regression approach to speech enhancement based on deep neural
  networks,''
\newblock {\em IEEE/ACM Transactions on Audio, Speech, and Language
  Processing}, vol. 23, no. 1, pp. 7--19, 2015.

\bibitem{xia2013speech}
Bingyin Xia and Changchun Bao,
\newblock ``Speech enhancement with weighted denoising auto-encoder.,''
\newblock in {\em INTERSPEECH}, 2013, pp. 3444--3448.

\bibitem{osako2015complex}
Keiichi Osako, Rita Singh, and Bhiksha Raj,
\newblock ``Complex recurrent neural networks for denoising speech signals,''
\newblock in {\em Applications of Signal Processing to Audio and Acoustics
  (WASPAA), 2015 IEEE Workshop on}. IEEE, 2015, pp. 1--5.

\bibitem{abdel2014convolutional}
Ossama Abdel-Hamid, Abdel-Rahman Mohamed, Hui Jiang, Li~Deng, Gerald Penn, and
  Dong Yu,
\newblock ``Convolutional neural networks for speech recognition,''
\newblock {\em IEEE/ACM Transactions on audio, speech, and language
  processing}, vol. 22, no. 10, pp. 1533--1545, 2014.

\bibitem{amodei2015deep}
Dario Amodei, Rishita Anubhai, Eric Battenberg, Carl Case, Jared Casper, Bryan
  Catanzaro, Jingdong Chen, Mike Chrzanowski, Adam Coates, Greg Diamos, et~al.,
\newblock ``Deep speech 2: End-to-end speech recognition in english and
  mandarin,''
\newblock {\em arXiv preprint arXiv:1512.02595}, 2015.

\bibitem{mao2016image}
Xiao-Jiao Mao, Chunhua Shen, and Yu-Bin Yang,
\newblock ``Image denoising using very deep fully convolutional encoder-decoder
  networks with symmetric skip connections,''
\newblock {\em arXiv preprint arXiv:1603.09056}, 2016.

\bibitem{he2015deep}
Kaiming He, Xiangyu Zhang, Shaoqing Ren, and Jian Sun,
\newblock ``Deep residual learning for image recognition,''
\newblock {\em arXiv preprint arXiv:1512.03385}, 2015.

\bibitem{vincent2010stacked}
Pascal Vincent, Hugo Larochelle, Isabelle Lajoie, Yoshua Bengio, and
  Pierre-Antoine Manzagol,
\newblock ``Stacked denoising autoencoders: Learning useful representations in
  a deep network with a local denoising criterion,''
\newblock {\em Journal of Machine Learning Research}, vol. 11, no. Dec, pp.
  3371--3408, 2010.

\bibitem{ioffe2015batch}
Sergey Ioffe and Christian Szegedy,
\newblock ``Batch normalization: Accelerating deep network training by reducing
  internal covariate shift,''
\newblock {\em arXiv preprint arXiv:1502.03167}, 2015.

\bibitem{nair2010rectified}
Vinod Nair and Geoffrey~E Hinton,
\newblock ``Rectified linear units improve restricted boltzmann machines,''
\newblock in {\em Proceedings of the 27th International Conference on Machine
  Learning (ICML-10)}, 2010, pp. 807--814.

\bibitem{garofolo1993darpa}
John~S Garofolo, Lori~F Lamel, William~M Fisher, Jonathon~G Fiscus, and David~S
  Pallett,
\newblock ``Darpa timit acoustic-phonetic continous speech corpus cd-rom. nist
  speech disc 1-1.1,''
\newblock {\em NASA STI/Recon technical report n}, vol. 93, 1993.

\bibitem{akkermans2011freesound}
Vincent Akkermans, Frederic Font, Jordi Funollet, Bram De~Jong, Gerard Roma,
  Stelios Togias, and Xavier Serra,
\newblock ``Freesound 2: An improved platform for sharing audio clips,''
\newblock in {\em Klapuri A, Leider C, editors. ISMIR 2011: Proceedings of the
  12th International Society for Music Information Retrieval Conference; 2011
  October 24-28; Miami, Florida (USA). Miami: University of Miami; 2011.}
  International Society for Music Information Retrieval (ISMIR), 2011.

\bibitem{mowlaee2013iterative}
Pejman Mowlaee and Rahim Saeidi,
\newblock ``Iterative closed-loop phase-aware single-channel speech
  enhancement,''
\newblock {\em IEEE Signal Processing Letters}, vol. 20, no. 12, pp.
  1235--1239, 2013.

\bibitem{glorot2010understanding}
Xavier Glorot and Yoshua Bengio,
\newblock ``Understanding the difficulty of training deep feedforward neural
  networks.,''
\newblock in {\em Aistats}, 2010, vol.~9, pp. 249--256.

\bibitem{le2015simple}
Quoc~V Le, Navdeep Jaitly, and Geoffrey~E Hinton,
\newblock ``A simple way to initialize recurrent networks of rectified linear
  units,''
\newblock {\em arXiv preprint arXiv:1504.00941}, 2015.

\bibitem{Adam}
Diederik~P. Kingma and Jimmy Ba,
\newblock ``Adam: {A} method for stochastic optimization,''
\newblock {\em CoRR}, vol. abs/1412.6980, 2014.

\bibitem{vincent2006performance}
Emmanuel Vincent, R{\'e}mi Gribonval, and C{\'e}dric F{\'e}votte,
\newblock ``Performance measurement in blind audio source separation,''
\newblock {\em IEEE transactions on audio, speech, and language processing},
  vol. 14, no. 4, pp. 1462--1469, 2006.

\bibitem{taal2010short}
Cees~H Taal, Richard~C Hendriks, Richard Heusdens, and Jesper Jensen,
\newblock ``A short-time objective intelligibility measure for time-frequency
  weighted noisy speech.,''
\newblock in {\em ICASSP}, 2010, pp. 4214--4217.

\bibitem{rix2001perceptual}
Antony~W Rix, John~G Beerends, Michael~P Hollier, and Andries~P Hekstra,
\newblock ``Perceptual evaluation of speech quality (pesq)-a new method for
  speech quality assessment of telephone networks and codecs,''
\newblock in {\em Acoustics, Speech, and Signal Processing, 2001.
  Proceedings.(ICASSP'01). 2001 IEEE International Conference on}. IEEE, 2001,
  vol.~2, pp. 749--752.

\end{thebibliography}

\end{document}